\documentclass[sn-mathphys]{sn-jnl}

\jyear{2021}%

\usepackage{amsmath}
\usepackage{amssymb}
\usepackage{booktabs}
\usepackage{graphicx}
\usepackage{subfigure}
\usepackage{times}
\usepackage{latexsym}
\usepackage[T1]{fontenc}
\usepackage[utf8]{inputenc}
\usepackage{microtype}
\usepackage{float}
\usepackage{indentfirst}
\usepackage{multirow}
\usepackage[justification=centering]{caption}

\theoremstyle{definition}

\theoremstyle{remark}

\numberwithin{equation}{section}

\begin{document}

\title[A Quadratic 0-1 Programming Approach for Word Sense Disambiguation]{A Quadratic 0-1 Programming Approach for Word Sense Disambiguation}


\author*[]{\fnm{Boliang} \sur{Lin}}\email{bllin@bjtu.edu.cn}

\affil*[]{\orgdiv{School of Traffic and Transportation}, \orgname{Beijing Jiaotong University}, \orgaddress{\city{Beijing}, \postcode{100044}, \country{China}}}


\abstract{Word Sense Disambiguation (WSD) is the task to determine the sense of an ambiguous word in a given context. Previous approaches for WSD have focused on supervised and knowledge-based methods, but inter-sense interactions patterns or regularities for disambiguation remain to be found. We argue the following cause as one of the major difficulties behind finding the right patterns: for a particular context, the intended senses of a sequence of ambiguous words are dependent on each other, so the choice of one word’s sense is associated with the choice of another word’s sense, making WSD a combinatorial optimization problem. In this work, we approach the interactions between senses of different target words by a Quadratic 0-1 Integer Programming model (QIP) that maximizes the objective function consisting of (1) the similarity between candidate senses of a target word and the word in a context (the sense-word similarity), and (2) the semantic interactions (relatedness) between senses of all words in the context (the sense-sense relatedness).}




\maketitle

\section{Introduction}
\label{sec:intro}

Word Sense Disambiguation (WSD) is the task to recognize the correct sense of a target word in a context. As one of the most difficult problems in AI, its challenges come from various sources: the power-law distribution between the more frequent and the less frequent words, the lack of labeled data for the less frequent senses, the fine granularity of sense inventories, the heavy reliance on knowledge, and whether there is a single target word per context or a sequence of words to disambiguate \cite{navigli}.

Over the past decades, supervised machine learning methods and knowledge-based methods have been the main horsepower to tackle WSD. The Lesk algorithm \cite{lesk} leverages the knowledge of sense definitions, and proposes to represent senses and target words as bags of words of sense definitions and contexts respectively and then assign the sense whose definition has the highest overlap with the context as the correct sense to the target word. WordNet \cite{wordnet1,wordnet2} is a common source of gloss. Supervised methods make use of sense-annotated data and train classification models, from building word expert classifiers
\cite{zhong} to exploiting deep learning architecture \cite{hcan,ewise}. With the recent success of pre-trained language models, contextualized embeddings have become the new representations for target words and senses. Incorporating BERT \cite{bert} into WSD has shown to be very effective to produce such embeddings and improve the all-words WSD performance \cite{glu,lmms,svc,glossbert,bem,sensembert,sref,ewiser,ares,csi}.

In particular, the comparison between a sense embedding and the contextual embedding of a target word, either through a dot product or distance-based metrics, has been at the heart of scoring each candidate sense of a target word in several state-of-the-art WSD systems \cite{bem,sref,sensembert,ares,csi}. For example, the bi-encoder model (BEM; \cite{bem}) jointly fine-tunes a BERT based context encoder and gloss encoder, and uses dot product to score the context embedding with each candidate sense embedding, produced by the respective encoders. Does this sense-word similarity best explain or serve the disambiguation objective? We argue that something could be missing here: based on the premise that words in the same context tend to share a common topic \cite{lesk}, the semantic relatedness between senses of all target words in a context could also affect the scoring and thus prediction.

In particular, our contributions are two-fold:

1. Unlike existing methods, we put in the all interactions between senses of all words in the context to form a quadratic 0-1 integer programming model, not only consider the interactions between neighbors words.

2. Different from the existing works, we propose to incorporate all interactions between senses of all words in the context, not only consider the interactions between neighbors words.

\section{Related Work}

Recently, \cite{game} solves WSD by a new model formulated in terms of evolutionary game theory, where each word to be disambiguated is represented as a node on a graph whose edges represent word relations and senses are represented as classes. However, these optimization approaches focus on the relatedness according to the senses that neighboring words. When WSD is viewed as a combinatorial optimization problem, the task is to find a sequence of senses that maximizes the semantic relatedness among all target words in a neighborhood \cite{pbp,dbees}. Though WSD can work with any measure that captures the relatedness between senses, the best measure for semantic relatedness has been a long-standing problem. Before the currently employed variant of comparing sense definition overlaps to a single target word context, the Lesk algorithm identifies the correct senses by finding the highest gloss overlap of all word senses in a context. Unfortunately, the need of exhaustive search prohibits the algorithm to work as the number of target words grow, and the bag-of-words gloss overlap as a measure of semantic relatedness is easily affected by the exact wording of a sense gloss. Later, \cite{pbp} adapts Lesk by considering gloss overlaps between two senses and also related concepts, and assigns a sense to a target word in an iterative manner. A wide array of relatedness measures have been evaluated, such as WordNet path and information content based metrics.

Different from the above, we propose to incorporate all interactions between senses of all words in the context, not only consider the interactions between neighbors words. To this end, we view WSD as a combinatorial problem, and aim to study the interference of the two terms.

\section{Semantic interactions Measures between senses}

\subsection{Drawbacks of Concept Network for WSD}
\begin{figure*}[htp]
\centering
\setlength{\abovecaptionskip}{-2pt}
\subfigure[Original order of disambiguation]{
\label{fig:1a}
\includegraphics[width=5.3cm,height=3cm]{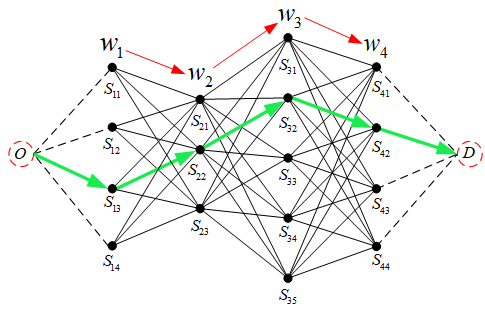}}
\subfigure[Changed order of disambiguation]{
\label{fig:1b}
\includegraphics[width=5.3cm,height=3cm]{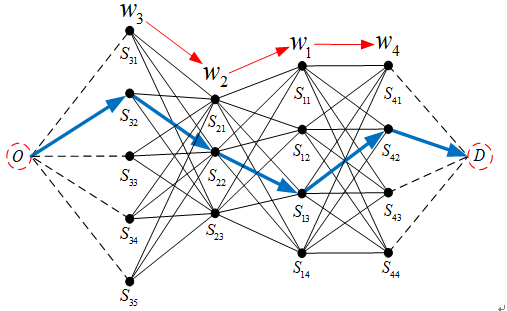}}
\caption{The influence of disambiguation order for concept networks of WSD}
\end{figure*}

By graph theory, the task of WSD can be viewed as finding paths in a concept network (Pedersen, Banerjee, and Patwardhan 2005). However, exact solutions to this pathfinding problem are hard to obtain by applying shortest or longest path algorithms. We will elaborate the difficulties with the following example of a concept network consisting of 4 target words. The target words are denoted by $w_1$,$w_2$,$w_3$ and $w_4$ following their order of appearance in a given context. $w_1$ has 4 candidate senses, denoted by $s_{11}$,$s_{12}$,$s_{13}$ and $s_{14}$, and $w_2$,$w_3$ and $w_4$ has 3, 4, 5 candidate senses respectively which are denoted similarly, then we obtain 16 possible senses.
In Figure \ref{fig:1a}, $w_1$,$w_2$,$w_3$ and $w_4$ are connected by the word order in a given sentence. Each sense associated with a word is represented by a node, and a node of a word is connected to the nodes of its neighboring words. We add a virtual origin node $O$ and a virtual destination node $D$ to the network. In this setting, the goal of WSD is to find an optimal path in the multilayer network, where there are 4×3×5×4 possible paths. For example, the green path is one of them: $O$$\rightarrow$$s_{11} $$\rightarrow$$s_{12}$$\rightarrow$$s_{13}$$\rightarrow$$s_{14}$$\rightarrow$$D$. Denote the set of all possible paths as $P$, then the pathfinding problem can be stated mathematically by

\begin{equation}
    l^*=argmax\{d_l:l\in P\}
\label{eq:1}
\end{equation}

where $d_l$ denotes the length of path $l$ (Note that Equation \ref{eq:1} can also be formulated as a minimization problem for generalized distance $d_l$). If each edge is fixed, the computational complexity for the problem is $O(n^3)$, where $n$ is the number of nodes. Unfortunately, $d_l$ can not be represented as a static value. For the green path, denote its length as $d^{'}$, then we have
\begin{equation}
\begin{aligned}
d^{'}&=c_{13}+h_{13,22}+c_{22}+h_{22,32}+c_{32}+h_{32,42}+c_{42}
\end{aligned}
\label{eq:2}
\end{equation}

where $c_{im}$ is the weight between a target word $w_i$ and its candidate sense $s_{im}$, and $h_{im,jn}$ is the weight of the edge between sense $s_{im}$ of word $w_i$ and sense $s_{jn}$ of word $w_j$.

The above concept network explicitly follows the word order in the sequence, but we consider that the disambiguation of a series of words does not have to be done following any specific order of words, in fact, when we switch the disambiguation order of any two words, changes will occur in the edges of the concept network and the resulting paths.

Figure \ref{fig:1b} illustrates what happens when we switch the disambiguation order to $O\rightarrow w_3 \rightarrow$ $w_2 \rightarrow w_1 \rightarrow w_4 \rightarrow D$. The length for the new green path is:
\begin{equation}
\begin{aligned}
d^{''}&=c_{32}+h_{32,22}+c_{22}+h_{22,13}+c_{13}+h_{13,42}+c_{42}
\end{aligned}
\label{eq:3}
\end{equation}

The edge connecting $s_{13}$ and $s_{42}$, thus the edge weight $h_{13,42}$ in Figure
\ref{fig:1b} do not exist in the concept network of Figure \ref{fig:1a}, and it generally does not equal to $h_{32,42}$, and hence $d^{'}\neq d^{''}$. Therefore, it is difficult to obtain an optimal solution for WSD by finding the shortest path in a single concept network. Considering the fact that the disambiguation order is not fixed, WSD is a combinatorial optimization problem at heart.

\subsection{Similarity and Semantic Relatedness Measures}\label{sec:32}
Earlier lexical similarity and semantic relatedness measures are often based on WordNet path lengths between two senses or information content. Recently, pre-trained BERT has helped improve WSD performance considerably. BEM uses two jointly fine-tuned BERT encoders to generate contextualized word embeddings and sense embeddings and compute dot products of the two embeddings to measure the similarity between a word and one of its candidate senses, which have shown to achieve good WSD performance. In detail, for a word in an input text, the context encoder will output the vector at the corresponding position as the word’s contextualized embedding; for a sense, the gloss encoder will take the WordNet sense gloss as input and output the vector for the [CLS] token as the sense embedding. We follow BEM and use their released model to generate word and sense embeddings for our experiments.

\begin{figure}[htbp]
\centering
\includegraphics[width=0.8\textwidth]{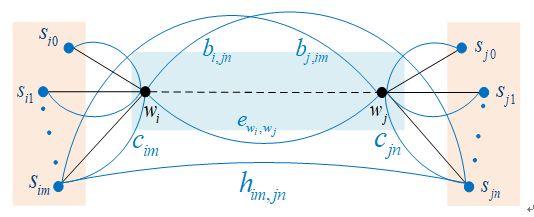}
\caption{The semantic network between two target words}
\label{fig:2}
\end{figure}

Extended from the embedding-based similarity measurements, we define the similarity and relatedness measures as follows, assuming that the contextualized word embeddings and sense embeddings are pre-trained, and cosine similarity is adopted for calculating the similarity measure
between two concepts. Let $S_i=\{s_{i1},s_{i2},...,s_{im},...,s_{iM}\}$ denote the candidate senses for word $w_i$, and $S_j=\{s_{j1},s_{j2},...,s_{jn},...,s_{jN}\}$ for word $w_j$. Figure
\ref{fig:2} shows a semantic network for the two words $w_i$ and $w_j$.

For the similarity measure, we redefine $c_{im}$ to be the similarity between the contextualized embedding of a target word $w_i$ and the sense embedding of its own candidate sense $s_{im}$, and define $h_{im,jn}$ as the similarity between sense embeddings of two words, i.e., sense $s_{im}$ and sense  $s_{jn}$, and also define $e_{w_iw_j}$ as the similarity between the contextualized embeddings of words $w_i$ and $w_j$.

To measure semantic relatedness between two senses, we denote the relatedness between sense $s_{im}$ (of word $w_i$) and sense $s_{jn}$ (of word $w_j$) as $r_{im,jn}$.Given the various ways of connecting two senses as illustrated in Figure \ref{fig:2}, we can compute relatedness directly or through a third party. In the simplest way, we use similarity for relatedness, i.e., let $r_{im,jn}=h_{im,jn}$. We observe that the relatedness between two senses can be conveyed through the target words, and denote the similarity between word and word $w_j$’s sense $s_{jn}$ as $b_{i,jn}$, then the relatedness can be expressed as $r_{im,jn}=b_{i,jn}+b_{j,im}$. To generalize from it, we devise the following relatedness measure:

\begin{align}
\begin{split}
   r_{im,jn}&=\lambda _1(b_{i,jn}+b_{j,im})+\lambda_2h_{im,jn}+\lambda_3(c_{im}+e_{w_iw_j}+c_{jn})
\end{split}
\label{eq:4}
\end{align}

For the values of $\lambda_1$, $\lambda_2$ and $\lambda_3$, we tune them according to experiments, and the values used for each experiment are listed in Section \ref{sec:32}. During computation for $b_{i,jn}$, $b_{j,im}$, $h_{im,jn}$, $e_{w_iw_j}$, if any term has a negative value, we will take the absolute value of that term to avoid cancellation of each other.

\section{Mathematical Formulation of the QIP Model for interactions Measures further enhance WSD}
\label{sec:math}
For a target word $w_i$, if it is associated with multiple senses in the sense inventory, we need to decide which sense is the best to fit in the given context. Therefore, it is natural to use binary selection variables to model each sense. We define:

\begin{equation}
\begin{aligned}
\begin{split}
x_{ik}=
\left\{
\begin{array}{cc}
     1 &\text{If word \textit{i} selects sense \textit{k}}\\
     0 &\text{otherwise}
\end{array}
\right.
\end{split}
\label{eq:5}
\end{aligned}
\end{equation}

When $x_{ik}=1$, we regard the similarity $c_{ik}$ between $w_i$ and its sense $s_{ik}$ as a gain for $w_i$. Then the total gain for $w_i$ is $\sum_{k \in S_i} c_{ik}x_{ik}$, where $S_i$ denotes the set of senses for $w_i$.Similarly, when word $w_i$ selects sense $s_{im}$ and $w_j$ selects $s_{jn}$, then we regard the relatedness $r_{im,jn}$ as a gain for the interaction between senses of $w_i$ and $w_j$, and the total gain for the two words is $\sum_{u \in S_i} \sum_{v \in S_i} r_{iu,jv}x_{iu}x_{jv}$. Our goal is to maximize the total gains for all words and senses.

Let the set of all target words in a context be denoted as $V$, and using above-mentioned notations, the WSD problem can now be written with a quadratic 0-1 programming formulation (QIP) as follows:

\begin{equation}
\begin{aligned}
    \text{Max} \quad Z(X)=\sum_{i \in V}\ \sum_{k \in S_i}\ c_{ik}x_{ik}+\beta\sum_{i,j \in V}\ \sum_{u \in S_i}\ \sum_{v \in S_i}\ r_{iu,jv}x_{iu}x_{jv}
\end{aligned}
\label{eq:6}
\end{equation}

\begin{equation}
\begin{aligned}
  \text{S.t.} \quad \sum_{k \in S_i}\ x_{ik}=1 \quad \forall i\in V
\label{eq:7}
\end{aligned}
\end{equation}

\begin{equation}
\begin{aligned}
(c_{i^{*}}-c_{i^{\#}}) / c_{i^{*}} - x_{i^*} < \theta\quad \forall i\in V
\label{eq:8}
\end{aligned}
\end{equation}

\begin{equation}
\begin{aligned}
x_{ik}\in \{0,1\} \quad \forall i\in V, \quad \forall k\in S_i
\label{eq:9}
\end{aligned}
\end{equation}

where $\beta$ is an adjustment coefficient.$c_{i^*}$ and $c_{i^\#}$ denote the largest and second largest similarity score between $w_i$ and its candidate senses, and $\theta$ is a threshold value between $0$ and $1$. In the QIP objective function, the first term computes the overall similarity between target words and candidate senses; and the second term calculates the overall relatedness from interactions between any two senses that come from different target words. Constraint (\ref{eq:7}) is a unique constraint that indicates that each word can select only one sense from its semantic library as the optimal sense.Constraint (\ref{eq:8}) is applied to enhance computational efficiency by keeping some variables to constant values. If $\theta=1$, it equals to not using this constraint. When $\theta=0$, it amounts to having an objective function without the second term, as $(c_{i^*}-c_{i^\#}) / c_{i^*}<1$  always holds. In general,$\theta\in (0,1)$, the closer theta is to 1, the more variables to solve.Constraint (9) enforces the binary constraint on decision variables.

If the sense interaction between adjacent words is considered, that is, if the word order is fixed, the interaction between senses can be considered according to the structure of Figure 1. In this case, the objective function can be reduced to:

\begin{equation}
\begin{aligned}
\text{Max} \quad Z(X)=\sum_{i \in V}\ \sum_{k \in S_i}\ c_{ik}x_{ik}+\beta\sum_{i,j \in V}\ \sum_{u \in S_i}\ \sum_{v \in S_i}\ r_{iu,(i+1)v}x_{iu}x_{(i+1)v}
\label{eq:10}
\end{aligned}
\end{equation}

When the interaction between senses is not accounted, i.e., $\beta$ = 0, the QIP objective function becomes:

\begin{equation}
\begin{aligned}
\text{Max} \quad Z(X)=\sum_{i \in V} \sum_{k \in S_i} c_{ik}x_{ik}
\label{eq:11}
\end{aligned}
\end{equation}
Since there is no relatedness calculation between target words, $Z(X)$ in Equation (\ref{eq:11}) can be degenerated into $Z(X)=\sum_{i \in V} Z(X_i^{*})$, where $Z(X_i^{*})$  is:
\begin{equation}
\begin{aligned}
Z(X_i^{*})=\text{Max} \quad \sum_{k \in S_i}\ c_{ik}x_{ik}
\label{eq:12}
\end{aligned}
\end{equation}

Considering that the constraint (\ref{eq:7}) ensures that a sense is assigned to a word exactly once, Equation (\ref{eq:12}) equals finding the maximal value of the similarity between a target word and its own candidate senses, i.e., it is simplified into QIP-r, stands for the case where the relatedness interactions term is not included in the QIP objective function:
\begin{equation}
\begin{aligned}
\text{QIP-r:}  \quad \quad \quad\quad Z(X_i^{*})=\underset{k\in S_i}{\text{Max}}\quad\{c_{ik}\}
\label{eq:13}
\end{aligned}
\end{equation}

Equation (\ref{eq:13}) is the widely used method in many works such as BEM, which is maximum relatedness disambiguation. It can be seen that the QIP model has already embedded this maximum relatedness disambiguation objective.

\begin{figure*}[htp]
\centering
\includegraphics[width=1\textwidth]{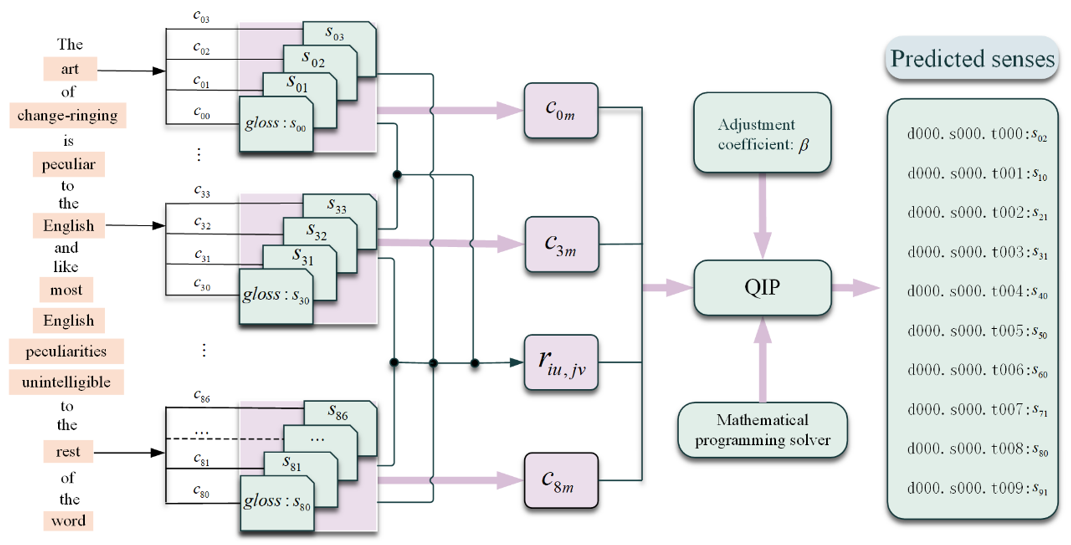}
\caption{The proposed QIP method for WSD}
\label{fig:3}
\end{figure*}

An overview of the QIP method is shown in Figure \ref{fig:3}.

\section{Conclusion}
In this work, we propose a new solution to WSD by constructing a quadratic 0-1 integer programming model. In fact, it is possible to implement WSD by selecting some good sense-sense interactions measures, i.e., we can easily determine which pre-trained vectors can be used as the sense-sense interactions.




\end{document}